\pdfoutput=1
\documentclass[11pt]{article}

\usepackage{acl}

\usepackage{times}
\usepackage{latexsym}
\usepackage[T1,T5]{fontenc}
\usepackage[vietnamese,english]{babel}
\usepackage[utf8]{inputenc}
\usepackage[T5]{fontenc}
\usepackage{microtype}
\usepackage{inconsolata}
\usepackage{graphicx}
\usepackage{booktabs}
\usepackage{multirow}
\usepackage{subcaption}
\usepackage{tabularx}
\usepackage{xcolor}
\usepackage{hyperref}

\title{VlogQA: Task, Dataset, and Baseline Models for Vietnamese Spoken-Based Machine Reading Comprehension}

  \author{Thinh Phuoc Ngo$^{1,2}$,
    Khoa Tran-Anh Dang$^{1,2}$, \\
    \textbf{Son T. Luu}$^{1,2}$, 
   \textbf{ Kiet Van Nguyen}$^{1,2}$,
    \textbf{Ngan Luu-Thuy Nguyen}$^{1,2}$ \\
    $^{1}$Faculty of Information Science and Engineering, University of Information Technology, \\Ho Chi Minh City, Vietnam\\
    $^{2}$Vietnam National University, Ho Chi Minh City, Vietnam\\
  \texttt{\{19520981,19520629\}@gm.uit.edu.vn, \{sonlt, kietnv, ngannlt\}@uit.edu.vn} \\}

\begin{document}
\maketitle
\begin{abstract}

This paper presents the development process of a Vietnamese spoken language corpus for machine reading comprehension (MRC) tasks and provides insights into the challenges and opportunities associated with using real-world data for machine reading comprehension tasks. The existing MRC corpora in Vietnamese mainly focus on formal written documents such as Wikipedia articles, online newspapers, or textbooks. In contrast, the VlogQA consists of 10,076 question-answer pairs based on 1,230 transcript documents sourced from YouTube -- an extensive source of user-uploaded content, covering the topics of food and travel. By capturing the spoken language of native Vietnamese speakers in natural settings, an obscure corner overlooked in Vietnamese research, the corpus provides a valuable resource for future research in reading comprehension tasks for the Vietnamese language. Regarding performance evaluation, our deep-learning models achieved the highest F1 score of 75.34\% on the test set, indicating significant progress in machine reading comprehension for Vietnamese spoken language data. In terms of EM, the highest score we accomplished is 53.97\%, which reflects the challenge in processing spoken-based content and highlights the need for further improvement.

\end{abstract}

\section{Introduction}
\label{intro}
Machine reading comprehension (MRC) is a natural language processing (NLP) task that requires machines to comprehend a given context to answer a question \cite{baradaran2022survey}. Although there are numerous datasets available for MRC tasks in English \cite{dzendzik-etal-2021-english}, existing datasets for reading comprehension tasks in Vietnamese are relatively limited and they have primarily focused on written documents, such as Wikipedia articles, textbooks, and online news articles.  Spoken language represents an important and distinct domain that has not been fully explored. Spoken language exhibits unique characteristics such as slang, regional variations, and informal grammar structures that can present significant challenges for machine learning models. As a result of that, reading comprehension tasks that involve spoken language, which is closer to everyday language, require a different type of dataset. 

To address this need, we introduce VlogQA - a new Vietnamese spoken language corpus for reading comprehension tasks originating from transcripts of YouTube vlogs. As a global online video-sharing and social media platform, YouTube provides a vast amount of spoken language data in natural settings. It is now the second-most visited website\footnote{https://www.similarweb.com/top-websites/} in the world (and Vietnam) and the second-biggest social network with over 2.5 billion monthly users\footnote{https://www.statista.com/statistics/272014/global-social-networks-ranked-by-number-of-users/}. Starting with YouTube, we aim to establish a solid foundation and gain insights into language patterns. This initial experiment serves as a stepping stone, and if successful, additional platforms catering to diverse audiences can be subsequently incorporated into further research. The dataset contains 10,076 manually annotated question-answer pairs based on 1,230 transcript documents extracted from YouTube videos. Besides, we provide several baseline models and evaluate them on our new dataset to test the ability of computers to understand the spoken text in Vietnamese. Overall, this paper makes the following contributions:
\begin{itemize}
    \item We introduce VlogQA, a new Vietnamese corpus for MRC tasks that focuses on natural spoken language. The corpus contains transcripts from videos covering the topics of food and travel and has a noticeably larger average transcript length compared to the context size of other similar datasets. The inclusion of spoken language data enhances the value of the corpus, making it an invaluable resource for research purposes. Additionally, this resource has the potential for developing and evaluating spoken language QA systems that leverage speech-to-text tools to extract information from recordings or live-stream videos. For instance, the corpus can facilitate the training of a QA system tailored for meeting recordings, thereby simplifying content extraction by obviating the need for extensive note-taking or traditional Meeting Minutes.
    
    \item We provide the creation process with sufficient annotation steps to assure the quality of the corpus. Besides, we conduct the analysis and comparisons regarding the corpus, including the number of question-answer pairs, length-based statistics, and the distribution of question types to get insight into the natural spoken language in Vietnamese. We choose UIT-ViQuAD - a pilot MRC corpus constructed on Vietnamese Wikipedia Texts, to perform a comprehensive comparative analysis for exploiting the characteristics of spoken language. 
    
    \item Finally, we evaluate the performance of multiple transformer-based language models on the corpus and analyze their performance for the MRC task on the spoken language domain. From the empirical results, we identify certain constraints within the dataset and highlight areas that can be improved in future studies.

\end{itemize}

The paper is structured as follows. Section 2 discusses existing studies. Section 3 is about corpus creation and its statistics. While Section 4 presents information about the language models to be used; the experimental results of human and language models, plus error analysis on the corpus are presented in Section 5. Finally, Section 6 provides conclusions and directions for future work. 

 \begin{table*}[ht]
    \centering
    \small
    \begin{tabularx}{\textwidth}{|X|>{\arraybackslash}p{0.5\textwidth}|p{0.18\textwidth}|}
        \hline
        \textbf{Question} & \textbf{Transcript} & \textbf{Answer} \\
        \hline
        \vspace{1mm}
      Nên chọn thịt như thế nào để không bị khô và vẫn giữ được độ mềm? \newline \textit{(\textcolor{blue!80!black}{What type of pork should be selected to avoid dryness while maintaining its softness?})} 
      & \vspace{1mm} […] \underline{thật} này mình sẽ xào cho nó \underline{chính} nhé thật này nó có \textbf{\textcolor{red}{vừa nạc vừa \underline{mõ}}} đó các bạn \underline{linh chi} Mần ăn thì nó sẽ có cái độ mềm mềm béo nhá chứ mình làm không mấy thì ăn nó rất khô […] \newline \textit{(\textcolor{blue!80!black}{we stir-fry the meat until it's really done the meat should be fatty meat type When being cooked it will have a tender texture otherwise it will be dry})}
      & \vspace{1mm} "answer\_start": 2196,  "text": "vừa nạc vừa mỡ" \textit{(\textcolor{blue!80!black}{fatty meat})} \\
    \hline
    \vspace{1mm}
    Vì sao điểm khảo cổ Sa Huỳnh phải đổi tên? \newline
    \textit{(\textcolor{blue!80!black}{Why did the Sa Huynh archaeological site have to change its name?})} 
    & \vspace{1mm}
    […]lần đầu được tìm thấy vào năm 1909 bởi nhà khảo cổ học người Pháp \underline{Venus} trước đây \underline{Nghĩa} danh này có tên là \underline{sao} Hoàng tức là \underline{nhạc} vàng song vì chữ hoàng lại \textbf{\textcolor{red}{trùng tên Với Chúa Nguyễn Hoàng}} cho nên đọc lái lại là thành sa huỳnh […] \newline \textit{(\textcolor{blue!80!black}{first discovered in 1909 by a France arrchaeologist, Vinet. the site once had a name Sa Hoang which means golden sand however Hoang is the same name as Lord Nguyen Hoang, so it had to be euphemized to Sa Huynh})} 
    
    & \vspace{1mm}
    "answer\_start": 893,  "text": "trùng tên Với Chúa Nguyễn Hoàng"  \textit{(\textcolor{blue!80!black}{(Hoang) is the same name as Lord Nguyen Hoang})}\\
    \hline
    
    \end{tabularx}
    \caption{\centering The examples in the corpus include ASR errors in Vietnamese, which are indicated by underlined text. The corresponding corrected English translations are also provided.}
    \label{fig:example}
\end{table*}
\section{Related Works}
\label{related}

UIT-ViQuAD \cite{UIT-ViQuAD} is a span-detection dataset for the Machine Reading Comprehension (MRC) task in Vietnamese, containing 23,074 questions on 5,109 passages acquired from Vietnamese Wikipedia articles. This dataset is widely used as a benchmark in Vietnamese MRC research and has facilitated innovations in the field. Its later version, UIT-ViQuAD 2.0 \cite{viquad2}, includes 9,217 additional unanswerable questions, which addresses a limitation of extractive MRC models that struggle to identify answers that are not explicitly mentioned in the text. Building upon the foundation of UIT-ViQuAD, UIT-ViWikiQA \cite{uitviwikiqa} is a sentence-detection dataset converted from UIT-ViQuAD and is designed for tasks that focus on sentence-level comprehension. In the health domain, ViNewsQA \cite{uitvinews} is a dataset comprising 22,057 questions on 4,416 online health articles from a popular newspaper in Vietnam.

Apart from span-detection datasets, there are other types of question-answering datasets available. ViMMRC \cite{vimmrc} is the first Vietnamese multiple-choice QA dataset, containing 2,783 four-choice questions based on 417 reading passages from Vietnamese literature textbooks. The second version of ViMMRC \cite{vimmrc2} introduces 699 reading passages and 5,273 questions with variable numbers of choices. UIT-ViCoV19QA \cite{uitvicov19qa} utilizes online FAQ documents from trusted healthcare organizations to address COVID-19-related questions, and is introduced as the first community-based QA dataset in Vietnamese with a total of 4500 questions. ViMQA \cite{vimqa} is a Wikipedia-based multi-hop dataset that provides over 10,000 questions designed to challenge models to perform complex multi-hop reasoning tasks, requiring them to refer to multiple evidence passages and perform explainable reasoning.

The availability and diversity of quality question-answering datasets are essential for the development of effective machine-learning models for natural language processing tasks. 
Spoken SQuAD \cite{spokensquad} is an English dataset that targets spoken content comprehension in the context of Wikipedia articles. It is derived from SQuAD \cite{squad} and employs text-to-speech tools to generate the spoken context. Similarly, the ODSQA \cite{odsqa} dataset focuses on spoken data and is based on the Delta Reading Comprehension Dataset (DRCD) \cite{drcd}, a Chinese contains 30.000+ questions from 2,108 Wikipedia articles. However, unlike Spoken SQuAD, ODSQA's audio is generated by humans.

 In summary, current Vietnamese MRC datasets have mainly concentrated on formal types of content, such as Wikipedia articles, textbooks, and online news articles. While there are spoken-based question-answering datasets available in other languages, such as Spoken SQuAD and ODSQA, they are still limited to Wikipedia content.

% Overall, the existing dataset for Vietnamese MRC research has primarily focused on written documents such as Wikipedia articles, textbooks, and online news articles. While these sources provide valuable information, they do not necessarily reflect the language used in everyday spoken interactions, which can be quite different in terms of vocabulary, grammar, and syntax.

\section{Corpus for Vlogs Reading Comprehension}
\label{corpus}
\vspace{-0.5cm}
\begin{center}
\begin{figure*}[ht!]
    \centering
    \includegraphics[width=\textwidth]{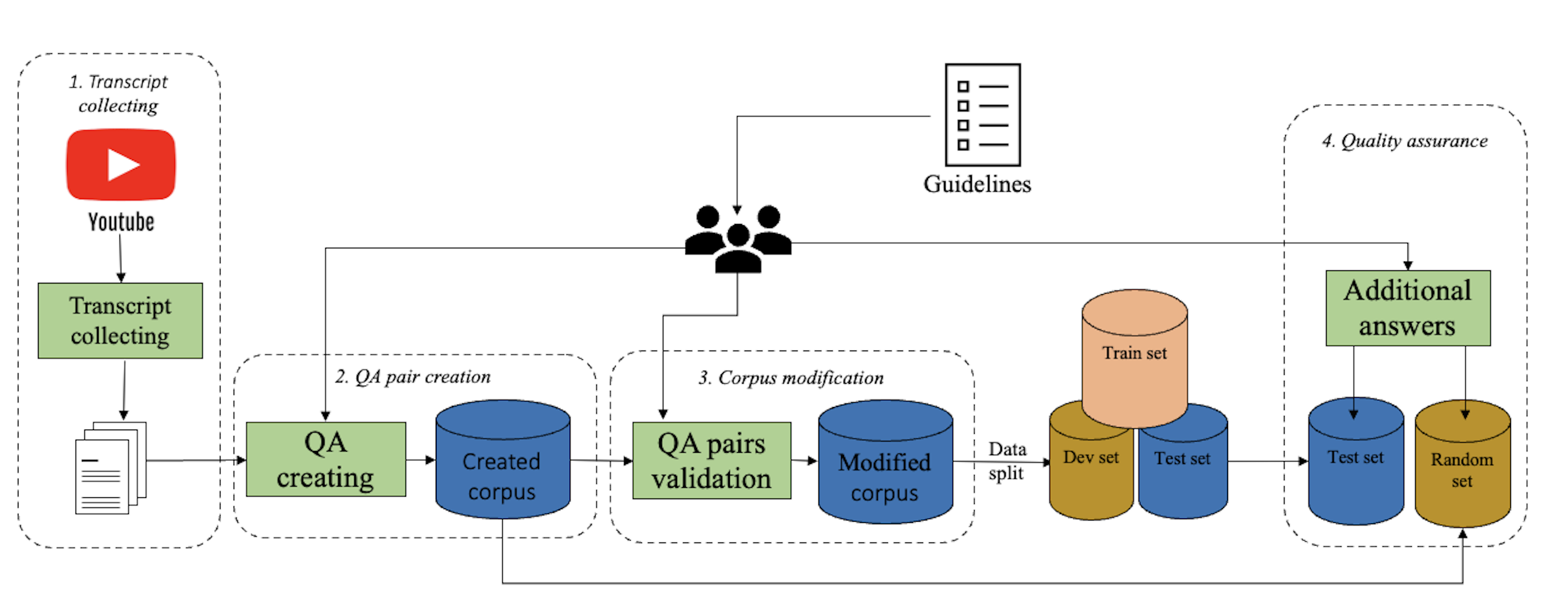}
    \caption{ The overview process of creating the corpus.}
    \label{fig:2}
\end{figure*}
\end{center}

    \subsection{Annotation Guidelines}
    Table \ref{fig:example} illustrates the structure of examples in the corpus, which is organized as a triplet \textit{(q, t, a)}. 
    We describe the reading-comprehension task in the scope of this paper as follows: Given a transcript document \textit{t} of a Youtube vlog, one must comprehend and extract the answer \textit{a} for the question \textit{q}. The answer \textit{a} must represent a specific word or phrase that is present in the transcript \textit{t}.
    
    % The first element, q, is a textual question, while the second element, t, refers to the transcript of a YouTube vlog. The third element, a, represents a specific word or phrase that must be present in the transcript t, and is used as the answer to the question q.
    
    Annotators play a vital role in ensuring the quality of the corpus by comprehending each transcript and creating at least five questions for it. If a transcript is too ambiguous or contains excessive ASR errors, annotators are advised to discard it. 
    Similar to other single span-detection MRC datasets, the answer to a given question must be derived from the transcript's context and represent the shortest continuous meaningful phrase that matches the question. In addition, the answer must be a whole word or phrase. It is recommended that annotators generate the questions using their own words and include a diverse range of question types, answers, and supporting evidence.

\subsection{Data Creation Process}
The proposed process for creating the VlogQA corpus includes four main stages: Transcript collection, QA pair creation, Corpus modification, and Quality assurance. Figure \ref{fig:2} illustrates the overview of the creation process for the corpus and the detailed description is provided as follows.

\subsubsection{Transcript collection}
The transcripts in the corpus were collected from Vietnamese YouTube vlogs with topics related to food and cooking tutorials, travel, or both. The channels that own the vlogs should have a large subscriber base; in this dataset, we set the minimum number of subscribers at 200,000 to ensure that the content is acceptable and relevant to a portion of the community. For each vlog, the transcript was collected using a Python API\footnote{https://pypi.org/project/youtube-transcript-api/} that returns a list of short speech-span transcriptions and is later combined into a single document. In this paper, the transcripts were kept in their original size and not segmented into smaller passages.

\subsubsection{QA pair creation}\label{qa-pairs}
Corresponding to each transcript document, one annotator is asked to read, comprehend and then create question-answer pairs following the annotation guidelines. Having completed this stage, the questions are collected and randomly chosen to form a set of 100 questions. This set is used to estimate the degree of agreement among annotators.

\subsubsection{Corpus modification}
To improve the consistency of the annotator and ensure corpus validity, the annotators are tasked with conducting the following 3 steps: 
(1) Additional training: annotators participate in additional training to better understand the evaluation criteria and guidelines. (2) Self-validation: annotators do a self-check of their own work to identify and correct any errors or inconsistencies, such as unclear questions, incorrect answers, lack of information questions, and incorrect boundary answers. (3) Cross-annotator-validation: the self-checked data is later reviewed by another annotator to ensure accuracy and consistency.
The modified dataset is divided into three subsets: train, test, and development with a ratio of 8:1:1 based on the question. Each transcript is assigned to only one subset.
% These steps are designed to minimize errors and inconsistencies in the annotation process and to improve the overall quality of the corpus.

% (1) undergoing additional training to better understand the evaluation criteria and guidelines, (2) conducting a self-check of their own work to identify and correct any errors or inconsistencies, including unclear questions, incorrect answers, lack of information questions, and incorrect boundary answers, and (3) engaging in a process, which involves cross-checking each other's work. 
% \begin{enumerate}
%     \item \textbf{Self-validation:} Annotators first re-read their created triplets. They are requested to modify or replace the triplets if there exists one of the following mistakes: unclear questions, incorrect answers, lack of information questions, and incorrect boundary answers.
%     \item \textbf{Cross-annotator-validation:} Another annotator afterward reviews the self-validated triplets. The task in this step is similar to self-validation.
% \end{enumerate}

\subsubsection{Quality assurance}
To determine the reliability of the corpus, we perform the following two examinations:

\begin{enumerate}
    \item \textbf{Inter-rater agreement:} 
This step aims to estimate the quality of the annotators' work. Each annotator independently provides an additional answer for each question in the random set. During the process, annotators work without referring to the corpus's answers.
To estimate the inter-rater agreement, a measure of the degree to which annotators agree on their labels, we employ three metrics: Cohen's Kappa \cite{cohen}, Fleiss' Kappa \cite{fleiss}, and Krippendorff's alpha \cite{krippendorff}. Additionally, we also calculate the overlap among answers using ROUGE metrics \cite{rouge}, and compute the semantic similarity of answers among annotators by using BERTScore \cite{zhang2019bertscore}. 
    
    \item \textbf{Human performance:} 
    After splitting the corpus, an independent team is enrolled to augment the test set with additional answers. The F1 score and exact-match metrics are used to evaluate human performance on the dataset.

\end{enumerate}

\subsection{Dataset Analysis}\label{analysis}

\subsubsection{Overall statistics}\label{overiew-analysis}

     \begin{table*}[ht]
    \centering
    \begin{tabular}{@{}lcccccccc@{}}
    \toprule
    & \multicolumn{4}{c}{\textbf{VlogQA}} & \multicolumn{4}{c}{\textbf{UIT-ViQuAD}} \\
        \cmidrule(lr){2-5}\cmidrule(lr){6-9}
            & \textbf{Train} & \textbf{Dev} & \textbf{Test} & \textbf{Total}  
            & \textbf{Train} & \textbf{Dev} & \textbf{Test} & \textbf{Total}\\ \midrule
    \textbf{Context count} & 945& 130& 155& 1,230 &137&18&18&174\\ %\midrule
    \textbf{Question count} & 8,047& 1,017& 1,012& 10,076 &18,579&2,285&2,210&23,074\\ \midrule
    \textbf{Avg. context length} & 2,789.5& 2,779.5& 2,498.9& 2,751.7 & 153.7 &148.8 &155.8 &153.4  \\ %\midrule
    \textbf{Avg. question length} & 10.09& 10.10& 10.00& 10.08  &11.23  &11.96 &12.29 & 11.40  \\ %\midrule
    \textbf{Avg. answer length} & 3.22& 3.27& 3.31& 3.24  &8.06   &8.45 &8.93 & 8.18     \\ %\midrule
    \textbf{Vocabulary size} & 34,288& 12,639& 13,336& 39,211 & 36,940&9,746&10,263&42,545      \\ \bottomrule
    \end{tabular}

    \caption{Overall statistics of our dataset and UIT-ViQuAD.}
    \label{tab:overview-analysis}
    \end{table*}
    
    Inferred from Table \ref{tab:overview-analysis}, the dataset comprises 1,230 vlog transcripts, of which only 64 transcripts are manually created by video creators; the remaining are generated automatically by Youtube.  As shown in Table \ref{tab:trans-len}, most of the transcripts have less than 5,000 words, the shortest transcript consists of 223 words while the longest one has 38,228 words.

   Vietnamese relies heavily on word order and function words to convey meaning and express grammatical relationships, rather than inflectional affixes. Words in Vietnamese are constructed from syllables ({\foreignlanguage{vietnamese}{"tiếng"}}), which are the basic unit of meaning, and words can be mono-syllabic or poly-syllabic. Vietnamese is also known for its extensive use of compound words, which combine two or more words to create a new word with a distinct meaning \cite{vietnamese}. 
   % For example, the word {\foreignlanguage{vietnamese}{"điện thoại di động"}} means "mobile phone" in English, and is made up of the words {\foreignlanguage{vietnamese}{"điện thoại"}} (phone) and {\foreignlanguage{vietnamese}{"di động"}} (mobile). The syllable {\foreignlanguage{vietnamese}{"điện"}} represents the concept of electricity, while {\foreignlanguage{vietnamese}{"thoại"}} represents speech or voice. When combined, they create a compound word that represents the concept of a phone or communication device.
    Segmentation is essential for identifying the tones of syllables in a word, which can affect the meaning of the word and the overall meaning of a sentence. However, the Vietnamese language lacks a standard for word segmentation \cite{standard}.
     We use a Python Vietnamese toolkit\footnote{https://pypi.org/project/pyvi/0.1.1/} to segment words, following the methodology of the UIT-ViQuAD paper. We also re-calculate some statistics of UIT-ViQuAD v1.0, using the latest version of the tool to compare the two datasets.

    In comparison to UIT-ViQuAD, which consists of 23,074 questions, our dataset is smaller, with 10,076 questions. On average, the length of the questions between the 2 datasets is not much different; however, our answers are significantly shorter, only 3.24, compared with 8.18 words per answer of UIT-ViQuAD. Our dataset used more context documents, a total of 1,230 transcripts compared with 174 passages. Additionally, the transcripts in our dataset are much longer on average, with an average length of 2,751.7 words, compared to the majority of UIT-ViQuAD's context passages ranging from 101 to 200 words. 
    
    Despite the difference in the number of questions, our dataset offers a vocabulary size of 39,211, which is only 7.83\% less than UIT-ViQuAD with a vocabulary size of 42,545. In this study, the vocabulary is estimated based on the segmented words of the context documents. Of the two corpora, there are 13,647 overlapping words, and our corpus has a unique vocabulary of 25,564 words. The most frequent words and phrases in our dataset are related to unit measurements, linking words, padding words, and pronouns. Those are commonly used in everyday scenarios and may be considered informal or unlikely to appear in formal writing or contexts. In Appendix \ref{vocab}, we provide further details on the differences in vocabulary between the two datasets and the methods we used to identify them using word clouds.

% \begin{center}
% \begin{figure}[ht]
%     \centering
%     \includegraphics[width=.5\textwidth]{images/hist_len.png}
%     \caption{Histogram of transcript length.}
%     \label{fig:3}
% \end{figure}
% \end{center}

\begin{table}[ht!]
    \centering
    \begin{tabular}{@{}llcc@{}}
        \toprule
        \textbf{Length} & \textbf{Count} & \textbf{Percentage} \\
        \midrule
        0 - 2,000        & 4,446           & 44.1               \\
        2,000 - 4,000     & 3,315           & 32.9               \\
        4,000 - 6,000     & 1,674           & 16.6               \\
        6,000 - 8,000     & 492            & 4.9                \\
        9,000 - 39,000    & 149            & 1.5     \\
        \bottomrule
    \end{tabular}
    \caption{Transcript length distribution.}
    \label{tab:trans-len}
\end{table}

\subsubsection{Duration-based analysis}
The following information on video length is calculated based on a total of 1,221 videos, as not all videos were available at the time of statistics.
The results in Table \ref{tab:overall-length} reveal that the average length of the selected videos is 1,272.95 seconds (21.2158 minutes). The shortest video lasts 60 seconds (1 minute), while the longest video has a duration of 19,190 seconds (5.331 hours). On average, travel-related videos have longer durations than food-related videos.
Table \ref{tab:length-range} further supports the finding that the majority of videos from food channels have a duration between 400 to 1,200 seconds, while the majority of traveling channel videos typically range from 1,000 to 3,000 seconds.

\begin{table}[ht]
  \begin{subtable}[ht]{\linewidth}
    \centering
    \begin{tabular}{lccc}
       \toprule
            & \textbf{Food} & \textbf{Travel} & \textbf{Total} \\
            \midrule
            \textbf{Video count}  & 565           & 665             & 1,230           \\
            \textbf{Avg. length} & 721.86       & 1,914.91        & 1,272.95       \\
            \textbf{Max. length}    & 3,040          & 19,190           & 19,190          \\
            \textbf{Min. length}    & 173           & 60              & 60         \\   
       \bottomrule
    \end{tabular}

    \caption{Video length statistics by category (in seconds).}
    \label{tab:overall-length}
  \end{subtable}
  % \hspace{0.1\linewidth}
  \begin{subtable}{\linewidth}
    \centering
    \begin{tabular}{cccc}
    \toprule
        & \textbf{Length} & \textbf{Count} & \textbf{Percentage}       \\
        \midrule
    
    \multirow{4}{*}{\textbf{Food}} & $<$400          & 61        & 9.28 \\   
                                     & 400 - 800     & 383       & 58.30  \\  
                                     & 800 - 1,200    & 160       & 24.35  \\   
                                     & $>$1,600        & 53        & 8.07  \\   
    \midrule
      \multirow{4}{*}{\textbf{Travel}} & $<$1,000          & 66        & 11.70 \\
                                        & 1,000 - 2,000     & 291       & 51.60 \\
                                        & 2,000 - 3,000     & 151       & 26.77 \\
                                        & $>$3,000     & 56       & 9.93 \\
   \bottomrule
    
    \end{tabular}
\caption{Distribution of the video length (in seconds).}
\label{tab:length-range}
  \end{subtable}
  \caption{Statistics of video duration.}
  \label{tab:tables}
\end{table}

\subsubsection{Inter-rater agreement}
After the first annotation round (Section \ref{qa-pairs}), we calculate the inter-rater agreement among six annotators on three metrics. However, given that Cohen's Kappa \cite{cohen} is designed for two annotators, we calculated the average degree of Cohen's Kappa agreement among all possible pairs of annotators. The average level of inter-rater agreement, as demonstrated by the results in Table \ref{tab:inter-rate}, is approximately 0.44. This level of agreement falls within the moderate range \cite{landis}.

The results of the ROUGE metrics \cite{rouge} in Table \ref{tab:inter-rate} are significantly higher than the agreement degrees, suggesting that mismatches were mainly due to non-essential terms rather than fundamental disagreement on the answer. Although the annotators had captured the context, it also highlights that the Corpus modification stage should focus on improving the consistency of the annotators to ensure the reliability and validity of the corpus. Besides, the BERTScore \cite{zhang2019bertscore} value shows that the answers among annotators are significantly similar, ensuring high agreement between annotators.

\begin{table}[ht]
    \small
    \begin{center}
        \begin{tabular}{@{}lll@{}}
        \toprule
        \textbf{Metric}  & \textbf{Score}\\
        \midrule
        Cohen's Kappa (average)   & 0.4393 \\
        Fleiss' Kappa        & 0.4387 \\
        Krippendoff’s Alpha & 0.4398\\
        \midrule 
        RougeL & 0.7672 \\
        Rouge1 & 0.7683 \\
        Rouge2 & 0.6776 \\
        \bottomrule
        BERTScore & 0.8867 \\
        \bottomrule
        \end{tabular}
        \caption{Inter-rater agreement degree.}\label{tab:inter-rate}%
    \end{center}
\end{table}

\subsubsection{Question type analysis}\label{question_type}
We categorize the questions into seven types, namely Who, What, When, Where, Why, How, and Others. Additionally, the How-type questions are further divided into two subtypes: quantity-related questions, which inquire about the amount or number of something, and quality/method-related questions, which focus on the characteristics or techniques involved.
The question labeling process is done manually because the diversity of question words in Vietnamese makes it hard to automate the process. For example, the English question word "when" can be translated into various Vietnamese question words, such as {\foreignlanguage{vietnamese}{"khi nào", "lúc nào", "bao giờ"}}, and others, depending on the context. The word "nào" occurs in many of these translations, but applying rule-based methods is difficult because "nào" can also mean "what/which" in other contexts.
According to the statistics presented in Table \ref{tab:question-type-dis}, the distribution of question types in our dataset is different from that of UIT-ViQuAD. Although the proportions of the "What" type questions are similar in both datasets at 47.82\% and 49.97\%, our dataset has a larger proportion of questions of "How" type at 32.57\%, compared to 9.09\% in UIT-ViQuAD. This distribution of question types reflects the characteristics of the data domain, that food and travel content deliver large information about the quantity and it is easier for annotators to create questions of that type.

% \begin{center}
% \begin{figure}[ht]
%     \includegraphics[width=.45\textwidth]{images/qs_plot.png}
%     \caption{Question types proportions.}
%     \label{fig:quesdis}
% \end{figure}
% \end{center}

\begin{table}[h]

\centering
\begin{tabular}{ccc}
\toprule
                & \textbf{UIT-ViQuAD (\%)} & \textbf{VlogQA (\%)} \\
                \midrule
\textbf{What}   & 49.97               & 47.92               \\
\textbf{How}    & 9.09                & 32.57*                 \\
\textbf{Why}    & 7.54               & 8.63                  \\
\textbf{Where}  & 5.64                & 5.25                  \\
\textbf{When}   & 8.96                & 3.35                  \\
\textbf{Who}    & 9.41                & 2.22                  \\
\textbf{Others} & 9.41                 & 0.07                 \\
\bottomrule
\end{tabular}
\caption{The proportions of question types in UIT-ViQuAD and VlogQA dataset. In the VlogQA dataset, the How-type is the sum of the How-quantity type (25.59\%) and the How-quality type (6.98\%), respectively.
 }
\label{tab:question-type-dis}
\end{table}

\section{Models for Reading Comprehension}
\label{model}

% \subsection{Models}

Transformer \cite{transformer} is a type of neural network architecture designed to process sequential data. In this paper, we carry out the MRC task and evaluate performance on the following group of transformer-based pre-trained language models:

\begin{itemize}
\item Multilingual language models, including (1) mBERT \cite{bert} -- an extension of BERT developed by Google, having been trained on over 100 languages, and (2) XLM-R \cite{xlmr} -- a Cross-lingual Model introduced by Facebook Research.

% \item Monolingual language models, including (3) ALBERT \cite{albert}, (4) Longformer \cite{longformer}, which are English models; and (5) PhoBERT \cite{phobert}, (6) BARTPho \cite{bartpho}, (7) ViT5 \cite{vit5} which are Vietnamese models.

\item Monolingual language models, including (3) PhoBERT \cite{phobert}, (4) BARTPho \cite{bartpho}, (5) ViT5 \cite{vit5} which are constructed on Vietnamese data.

\end{itemize}

The information about the size of pre-trained language models, the hyperparameter settings, and the environment for experiments are shown in Appendix \ref{model_settings}.

% \begin{itemize}
%     \item mBERT, or Multilingual BERT \cite{bert}, is an extension of BERT developed by Google. Having been trained on a corpus from over 100 languages, it obtains state-of-the-art results on multiple NLP tasks.
%     \item XLM-R \cite{xlmr} is a Cross-lingual Language Model introduced by Facebook that has outperformed m-BERT on a variety of cross-lingual benchmarks.
%     \item ALBERT \cite{albert}, which stands for A Lite BERT, is a more compact version of BERT, a monolingual (English) pre-trained model. ALBERT achieves this efficiency by employing parameter-sharing techniques, which reduce the number of model parameters while maintaining or improving performance on a range of NLP tasks.
% \end{itemize}

\section{Empirical Results}
\label{results}

\subsection{Experimental results}

In this section, we first present the experimental results of the language models and compare their performance with that of humans. The models are fine-tuned using the training and development sets.

%---
% \setlength\tabcolsep{1em}
\begin{table}[h]

    \small
    \begin{tabular}{@{}lcccc@{}}
    \toprule
    \multirow{2}{*}{\textbf{Model}} & \multicolumn{2}{c}{\textbf{Dev (\%)}} & \multicolumn{2}{c}{\textbf{Test (\%)}} \\
        \cmidrule(lr){2-3} \cmidrule(lr){4-5}
            & \textbf{EM} & \textbf{F1}   
            & \textbf{EM} & \textbf{F1}\\ 
    \midrule
    \textbf{mBERT} &40.36 &61.60 &45.17 &64.89 \\ 
    \textbf{XLM-R$_{Base}$} &45.78 &65.63 &47.73 &68.71 \\
    \textbf{XLM-R$_{Large}$} &\textbf{51.41} &\textbf{72.39} & \textbf{53.97} &\textbf{75.34} \\
    % \textbf{ALBERT-V2} & 0.2771 & 0.4862 & 0.2652 & 0.4720  \\ 
    % \textbf{Longformer} & 0.2372 & 0.4023 & 0.2457 & 0.3973  \\ 
    \textbf{PhoBERT} & 23.32 & 35.49 & 23.06 & 34.37 \\
    \textbf{BARTPho} & 18.91 & 30.27 & 20.43 &  32.58 \\
    \textbf{ViT5} & 30.33 & 45.82 & 29.37 &  45.53 \\
    \midrule
    \textbf{Human Performance} & - & - & 48.49  & 76.25  \\
    \bottomrule
    \end{tabular}

    \caption{Pre-trained language models performance on VlogQA test set in terms of EM and F1-score. The models were trained on the VlogQA corpus.}
    \label{tab:models}
\end{table}

%---

The results in Table \ref{tab:models} indicate that the XLM-R$_{Large}$ model outperforms the other models, achieving the highest scores in both EM (53.97\%) and F1-score (75.34\%). In contrast, PhoBERT's performance was even lower than that of the English pre-trained models, possibly due to its word tokenizing technique, which may not be optimal for handling the challenges posed by spoken language. Spoken language often contains various errors, stutters, and other linguistic features unique to spoken communication, making it a challenging task for natural language processing models like PhoBERT. The ViT5 is the best performance monolingual pre-trained on the VlogQA dataset, which is 29.37\% for EM and 45.53\% for F1-score. However, the results of ViT5 is not as good as XLM-R on the VlogQA dataset. 

In order to assess human performance on the task, we computed the scores of two independent annotators on the test set. The resulting evaluation shows that the human performance achieved an EM score of 48.49\% and an F1-score of 76.25\%. Interestingly, the XLM-R$_{Large}$ model performed even better than humans on the EM metric, which is a remarkable accomplishment. However, on the F1-score, there is only a slight difference between the model and human performance. These findings suggest that the XLM-R$_{Large}$ model has the potential for this task, but there is still room for improvement in terms of F1-score.

\begin{table}[ht]
    \small
    \begin{center}
    
    \begin{tabular}{@{}lcc@{}}
        \toprule
        \textbf{Model}  & \textbf{EM (\%)}  & \textbf{F1-score (\%)}\\
        \midrule
        mBERT   & 16.67 & 37.05 \\
        XLM-R$_{Base}$      & 24.62 & 49.07 \\
        XLM-R$_{Large}$ &\textbf{35.42} & \textbf{62.43} \\
        % ALBERT-V2  & 0.0823 & 0.2594 \\
        % Longformer & 0.0301 & 0.1079 \\
        PhoBERT & 24.15 & 50.06 \\
        ViT5 & 8.28 & 19.30 \\ 
        BARTPho & 2.07 & 12.21 \\
        \bottomrule
        \end{tabular}
        \caption{Pre-trained language models performance on VlogQA test set in terms of EM and F1-score. The models were trained on the UIT-ViQuAD corpus.}
        \label{tab:test-viquad}
        \end{center}
\end{table}

In addition to training transformer-based models on our training and dev sets, we also evaluate the performance exclusively on the UIT-ViQuAD training and dev sets and then test them on our test set. This is to evaluate whether the current pre-trained models are good when they were fine-tuned on another domain. The results of our evaluation, as shown in Table \ref{tab:models} and \ref{tab:test-viquad}, indicate that XLM-R$_{Large}$ performs the best, but its performance decreases drastically when trained on the UIT-ViQuAD dataset. Surprisingly, the Vietnamese pre-trained model, PhoBERT, performs better when we train them on the UIT-ViQuAD dataset. However, it is still lower than the performance of the XLM-R. In general, the performance of the language model that was fine-tuned on the UIT-ViQuAD does not achieve the expected results for the MRC task on spoken text as it was fine-tuned on our VlogQA. 

\subsection{Error analysis}

\begin{table*}[!htp]
    \small
    \centering
    % \rotatebox{90}{
    \resizebox{.78\textwidth}{!}{
    \begin{tabular}{@{}lccccc@{}}
    \toprule
    \textbf{Question type}   &  \textbf{mBERT} &\textbf{XLM-R$_{Large}$ } & \textbf{PhoBERT}    & \textbf{BARTPho}  & \textbf{ViT5}\\
    \midrule
    What                  & 308  & 248 &387 &408 &353 \\
                      & 0.6123 &0.4930 &0.7694 &0.8111 &0.7018 \\
                    \midrule
    How (quantity)         & 97 & 80 &145 &159 &130 \\
                      &  0.4236 & 0.3493 &0.6332 &0.6943 &0.5677 \\
                    \midrule
    How (method)           & 56  & 49 &65 &65 &58\\
                           & 0.7887 & 0.6901   &0.9155 &0.9155 &0.8169 \\
                    \midrule
    Where                & 33  & 18 &39 &38 &35\\
                         & 0.6226 & 0.3396 &0.7358  &0.7170     &0.6604\\
                    \midrule
    Who                  & 18  & 12 &23 &23 &18 \\
                          & 0.6923  & 0.4615 &0.8846 &0.8846    &0.6923 \\
                    \midrule
    Why                 & 70 & 62  &76 &78 &74  \\
                        & 0.8434  & 0.7470 &0.9157 &0.9398 &0.8916 \\
                    \midrule
    When                & 18   & 16 &28 &32  &25 \\
                        & 0.5455  & 0.4848 &0.8485 &0.9697 &0.7576\\
                     \bottomrule
    
    \end{tabular}
    }
    % }
    \caption{The number and the rate of incorrect answers on the VlogQA development set, grouped by type, using the EM metric.}
    \label{tab:error}
\end{table*}

We exclude the "Others" question type in this section due to its negligible representation. Illustrated in Table \ref{tab:error} are the numbers of incorrect answers of each type and their proportions in the development set. An answer provided by the language model is considered wrong if the answer and the reference answer are not an exact match (${EM = 0}$). 
Overall, the XLM-R$_{Large}$ model achieves superior performance compared to other models in all question types. Therefore, we will focus on analyzing the errors of the XLM-R$_{Large}$ model.

Based on the information in Table \ref{tab:error}, the XLM-R model has the lowest error rate on Where and How (quantity) question types, at 33.96\% and 34.93\%. 
The What type questions make up the largest proportion of our dataset, giving an error rate of 49.30\%. The type that has the most error rate is the Why question type, with a rate of 74.70\%.

The average F1 score of error predictions is 43.18\%, and 73.46\% of predictions have a non-zero F1 score. Common errors can be categorized as inconsistent identification of non-essential terms, which may result from the variability of the spoken language. Other common errors include misinterpretation of the nuances of the question, and providing completely wrong answers that are not supported by the information provided. We further provide examples of the errors in Appendix \ref{error_example}.

\section{Conclusion and Future Works}
\label{conclusion}

This paper presents VlogQA - a new Vietnamese reading comprehension corpus for spoken context. The corpus consists of 10,076 question-answer pairs generated by humans, sourced from 1,230 transcripts of YouTube vlogs. Each transcript has an average length of 2,752 words. 
In terms of question types, the dataset is predominantly composed of What-questions, accounting for 47.52\% of the corpus. This is followed by How-questions, which make up 32.57\% of the dataset. Other question types represented in the corpus include When, Who, and Why, among others. Our experimental results indicate that the annotation of the dataset is acceptably consistent, with an average inter-rater agreement of nearly 44\%. The performance of the state-of-the-art multilingual model is comparable with humans in both F1-score and EM metrics; however, we believe that there is still room for improvement.
In future work, we plan to enhance the corpus both in quality and quantity. We will explore techniques for improving the consistency of annotations and seek to expand the dataset with additional transcripts, spanning more topics. We also plan to augment the corpus with unanswerable questions, which will enable further exploration of machine capabilities.

Overall, this new Vietnamese reading comprehension corpus for spoken context provides a valuable resource for researchers and practitioners in the field of natural language processing. Moreover, We anticipate this dataset will facilitate advancements in Vietnamese language understanding and provide a benchmark for the evaluation of intelligent question-answering systems on human-spoken language. Furthermore, this corpus will enable the development of smart systems capable of retrieving valuable information from spoken language, thus contributing to the advancement of the field.
\section*{Limitations}
\label{limitations}
% On one hand,
Using spoken content as a data source ensures that the corpus reflects the diverse nature of spoken language and culture of everyday life, including informal settings.
% Spoken language is a distinct domain posing unique challenges for machine learning models. 
On the other hand, these distinct resources of Youtube also pose unique challenges for existing systems, including:  
        \begin{itemize}
            \item 
        
             \textbf{Accent and dialect:} 
             % The videos were made by Vietnamese speakers who are native to the language. 
             Vietnamese is a tonal language with three main dialect regions (Northern, Central, and Southern), which means that there are differences in the way words are pronounced and used. This complexity and variation in real-life situations cause errors in automatic speech recognition (ASR) systems.
            \item
            \textbf{Audio quality:} Low-quality audio is difficult to transcribe accurately, leading to errors and inconsistencies in the dataset. Background noise, such as music or ambient sounds may interfere with the transcript quality, especially in outdoor recordings like travel vlogs.
            % \item
            % \textbf{Speaker identification:} Identifying speakers in dialogues or interviews is a challenging task, especially when multiple speakers are present. Unfortunately, YouTube's ASR system does not support this feature.            
            \item 
            \textbf{Transcript format:} Unlike regular documents, the ASR system does not provide punctuation (e.g., commas and periods) or consistent letter cases (e.g., uppercase first letters in named entities), which may pose challenges for understanding the meaning of the transcript. Moreover, ASR transcripts do not support identifying speakers where multi-speakers are present.            
            \item
            \textbf{Transcript length:} 
            The length of vlogs in our dataset is highly variable, with some videos lasting under 10 minutes and others exceeding an hour, leading to the fact that most of the transcripts are significantly larger than the context provided by other datasets.
           % Our dataset has an average of 2,752 words per transcript, but only 8.2 questions per transcript.
           There is a substantial amount of non-relevant information that needs to be filtered out to identify relevant information for each question.  
           % \item \textbf{Information density:}
           % Given the level of information, developing systems that can accurately identify and extract relevant information for the questions is a significant challenge.

        \end{itemize}
        
These factors may put a negative impact on MRC systems. However, they also present opportunities to provide a unique dataset with vocabulary and word combinations specific to spoken language, which is rare in the existing datasets. Finally, we could not make our own pre-trained language model on spoken language text due to the limitation of computing resources such as GPU and memory. We hope the future pre-trained language models and large language models (LLMs) for spoken texts will improve the performance of the machine reading comprehension model for spoken language.
\section*{Ethics Statement}
We select videos that are published and verified by YouTube, 94.80\% of the transcript documents are automatically generated by YouTube's speech recognition. We keep all selected transcripts in their original form, and they are available at the time of collection. For the data annotation process, all annotators are supported with adequate remuneration for their work. The information about annotators is made anonymous.

\section*{Acknowledgments}
This research was supported by The VNUHCM-University of Information Technology's Scientific Research Support Fund

\bibliography{custom}

\appendix
\onecolumn
\section{Appendices}
\subsection{Question type distribution}
\begin{figure*}[ht]
  \centering
  \begin{subfigure}[b]{\textwidth}
    \includegraphics[width=0.7\textwidth]{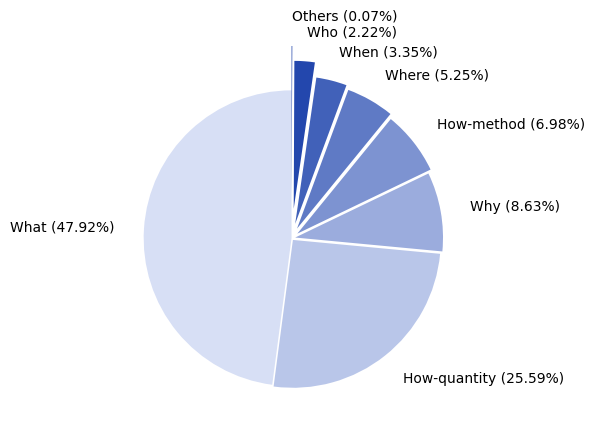}
    \caption{VlogQA}
    \label{fig:quesdis}
  \end{subfigure}
  \begin{subfigure}[b]{\textwidth}
    \includegraphics[width=.7\textwidth]{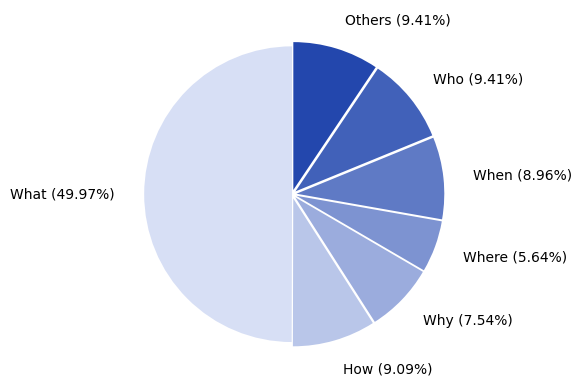}
    \caption{UIT-ViQuAD}
    \label{fig:uitquesdis}
  \end{subfigure}
  \caption{Question types proportions of the VlogQA and UIT-ViQuAD}
  \label{fig:quesdis2}
\end{figure*}

Figure \ref{fig:quesdis2} presents pie charts showing the distribution of question types in the VlogQA and UIT-ViQuAD, in addition to Section \ref{question_type}.

\subsection{Vocabulary}
\label{vocab}
\begin{figure*}[!h]
    \centering
    \begin{subfigure}[b]{0.48\textwidth}
        \includegraphics[width=\textwidth]{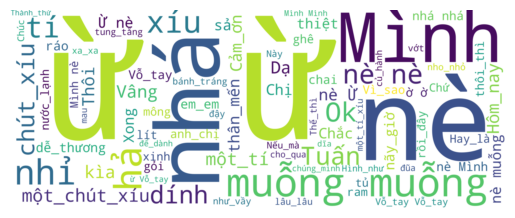}
        \caption{Word cloud of exclusive (tokenized) words in VlogQA from intersection vocabulary.}
        \label{fig:vocab1}
    \end{subfigure}
    \hfill
    \begin{subfigure}[b]{0.48\textwidth}
        \includegraphics[width=\textwidth]{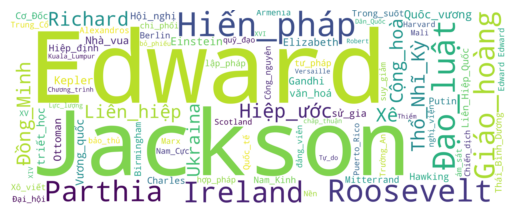}
        \caption{Word cloud of exclusive (tokenized) words in UIT-ViQuAD from intersection vocabulary.}
        \label{fig:vocab2}
    \end{subfigure}
    \begin{subfigure}[b]{0.48\textwidth}
        \includegraphics[width=\textwidth]{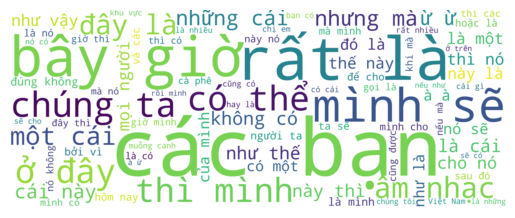}
        \caption{Word cloud of VlogQA (without tokenization).}
        \label{fig:vocab3}
    \end{subfigure}
    \hfill
    \begin{subfigure}[b]{0.48\textwidth}
        \includegraphics[width=\textwidth]{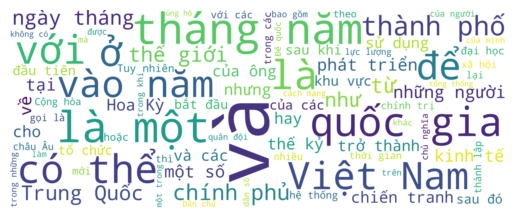}
        \caption{Word cloud of UIT-ViQuAD (without tokenization).}
        \label{fig:vocab4}
    \end{subfigure}
    \caption{Word cloud presentations of VlogQA and UIT-ViQuAD.}
    \label{fig:clouds}
\end{figure*}

Figure \ref{fig:clouds} shows the word clouds for the context documents in our dataset and UIT-ViQuAD. Each cloud is limited to 80 words, and we have opted not to apply any stop-word filters in these visualizations to preserve the essence of spoken materials.

As discussed in Section \ref{overiew-analysis}, our analysis revealed that the two corpora share an estimated 13,647 words. To further explore the distinctive vocabulary of each corpus, we created Figure \ref{fig:vocab1} to display a visualization of the exclusive vocabulary in our corpus, which does not overlap with the shared vocabulary.

The most frequent words in our exclusive cloud pertain to padding words in real-life spoken language that are eliminated in written contexts, such as "ừ", "nhé", and "nè". It also includes the pronoun "mình", which is common in informal contexts and is similar to "I", "we", "me", and "us" in English. Similarly, Figure \ref{fig:vocab2} presents a word cloud that showcases the unique words in UIT-ViQuAD. To generate these clouds, we tokenized the context documents to involve the estimated vocabulary.

The remaining part of Figure \ref{fig:clouds} presents the word clouds of the entire context documents of the two corpora. The UIT-ViQuAD cloud represents more formal words that frequently occur in informative contexts, such as "chính phủ", "tháng năm", "quốc gia", and "Việt Nam". On the other hand, VlogQA introduces a set of spoken language words, such as "rất là" to express emphasis on an adjective or adverb, or "các bạn", which is somewhat equivalent to "you guys" in English.

\subsection{Experimental settings}
\label{model_settings}
We train our baseline models in a Google Colab environment with a single NVIDIA Tesla T4 GPU.
The pre-trained models are fine-tuned using our dataset under the default settings of HuggingFace Trainer API\footnote{https://huggingface.co/docs/transformers/}, \textit{batch\_size} = 8 and \textit{epochs} = 10. We also set the \textit{max\_length} of the tokenizer to 512 (except for the case of PhoBERT, which is 256 due to the hardware limitations). The number of parameters for each model is described in Table \ref{tbl_num_param}. The baseline code is available at \url{https://github.com/sonlam1102/vlogqa}.

\begin{table}[ht]
    \centering
    \begin{tabular}{lc}
    \textbf{Model} & \textbf{\#parameters} \\
    \hline
    mBERT          & 179M                  \\
    XLM-R (base)   & 279M                  \\
    XLM-R (large)  & 561M                  \\
    PhoBERT        & 135M                  \\
    BARTPho        & 132M                  \\
    ViT5           & 310M                 \\
    \hline
    \end{tabular}
    \caption{Number of parameters for empirical models}
    \label{tbl_num_param}
\end{table}

\subsection{Incorrect prediction examples}
\label{error_example}
Table \ref{fig:error} illustrates a sample of incorrect predictions for each question type, derived from the XLM-R$_{Large}$ model on the VlogQA development set. The observed discrepancy between the model's F1 score and EM performance may be attributed to the extraneous or insufficient usage of non-essential terms, which is why we selected these particular examples.

For the What-type question, the model's prediction may include an optional excessive term, "region", but eliminating it does not significantly impact the results in the Vietnamese aspect. 

In the Quantity-type question, the term "dưới" (which means "under" in English) must be included in the answer to determine the constraint on the length of hair of Phu Quoc dogs. While the model's answer may be contextually relevant, it is not entirely accurate without the inclusion of the term "dưới".

In the case of the How-quality question, the model eliminated a term, which resulted in a minor grammatical flaw in Vietnamese (our translation to English may be deficient to fully reflect this example as the difference in the position of words in the 2 languages). It is worth noting that the model is extractive for spoken-based language, and in some contexts within this dataset, such a prediction/answer may still be acceptable despite the small error. However, our translation to English may also be deficient to fully reflect this example as the difference in the position of words in the two languages could affect the accuracy of the translation.

In the Why-type question, the model's prediction has correctly identified the context but may be missing some minor terms, as seen in previous examples. This elimination makes it more likely to be an answer to a What-type question in Vietnamese.

For the Where-type question, the model's prediction includes redundant terms that are not relevant to the question, which only seeks information about the location, not the designated name of it, the land of martial arts. While the model's answer may be partly correct, the inclusion of these unnecessary terms could potentially confuse the reader or listener and detract from the accuracy of the answer.

In the Who-type question, the model should not include terms that express an extreme as the question does not focus on this. The model's prediction should only include information that is relevant to answering the question and avoid adding unnecessary or extraneous information.

The illustrative examples provided do reflect the difficulties in processing Vietnamese spoken-based materials, particularly due to the complex grammar system and variations in pronunciation, intonation, and word order. While the questions in this dataset may not be considered hard, they can still be challenging for natural language processing models to accurately interpret and respond to. It is important to carefully consider the limitations of these models and the context in which they are being used when analyzing their performance on language tasks in Vietnamese.

\begin{table}[h!]

% \footnotesize
\begin{tabularx}{\textwidth}{|X|>{\centering\arraybackslash}p{0.18\textwidth}|p{0.45\textwidth}|p{0.1\textwidth}|p{0.1\textwidth}|}
    \hline
     & \textbf{Question} & \textbf{Transcript} & \textbf{Reference} & \textbf{Prediction} \\
    \hline
    \vspace{1mm}
\multirow{2}{*}{\rotatebox[origin=c]{90}{\textbf{What}}} &
  Khu vực Bắc Ninh từng được được gọi bằng tên gọi gì? &
  những cái khu chợ xưa ấy thì xung quanh đây là những cái cột đá cổ Nhưng cái xà thanh xà bằng gỗ và mà \underline{ấy máy} là mái ngói mái ngói theo đúng kiểu đặc trưng của vùng Bắc Bộ ngày xưa khu Bắc Ninh là gọi là \textbf{\textcolor{red}{khu} \textcolor{blue}{Kinh Bắc}} nhé \vspace{1mm} &
  Kinh Bắc & 
  khu Kinh Bắc \\
 &
  \textcolor{blue!80!black}{\textit{What was Bac Ninh called in the past?}} &
\textcolor{blue!80!black}{\textit{The old marketplaces had these ancient stone pillars around them. The wooden rafters and tiled roofs were of the typical style of the (classic Vietnamese) North. In the past, the region (around) Bac Ninh was called \textbf{\textcolor{blue}{Kinh Bac} \textcolor{red}{region}}}} &
  \textcolor{blue!80!black}{\textit{Kinh Bac}} &
 \textcolor{blue!80!black}{\textit{Kinh Bac region}} \\ \cline{1-5}
  \vspace{1mm}
\multirow{2}{*}{\rotatebox[origin=c]{90}{\textbf{How (quantity)}}} &
  Lông chó Phú Quốc dài bao nhiêu? &
  thứ nhất là nó phải lông \textcolor{blue}{\textbf{dưới 2cm}} cái Lông   nó sát gọi nó là \underline{quan} sát cái thứ 2 là nó cái anh em của nó và nó gọi cái   giới khoa học \underline{Tao} gọi là nó có \underline{cách} mạng bàn chân phát triển như là chân vịt \vspace{1mm} &
  dưới   2cm &
  2cm \\
 &
  \textcolor{blue!80!black}{\textit{What is the length of Phu Quoc dog's hair (in general)?}} &
  \textcolor{blue!80!black}{\textit{Firstly,   their hair must be \textbf{\textcolor{blue}{under 2cm}} which means it is \underline{short} Secondly, the   researchers observed that their feet have developed webbing similar to that   of ducks.}} &
  \textcolor{blue!80!black}{\textit{under   2cm}} &
  \textcolor{blue!80!black}{\textit{2cm }}\\ \cline{1-5}
  \vspace{1mm}
\multirow{2}{*}{\rotatebox[origin=c]{90}{\textbf{How (method)}}} &
  Tỉnh Bắc Ninh có ý nghĩa như thế nào với Hà Nội? &
  Tuy   nhiên không vì không vì mà nhỏ quá mà Bắc Ninh lại kém phát triển \underline{của} anh ạ   Bắc Ninh được coi \textbf{\textcolor{blue}{là thành phố vệ tinh của Hà Nội}} vì vậy là có rất nhiều   những cái khu công nghiệp đem lại giá trị kinh tế cao cho Việt Nam giống như   là Samsung \vspace{1mm} &
  là   thành phố vệ tinh &
  thành   phố vệ tinh \\
 &
  \textcolor{blue!80!black}{\textit{How important is Bac Ninh to Hanoi?}} &
  \textcolor{blue!80!black}{\textit{The fact that   Bac Ninh is small does not mean it is less developed, \underline{my} friend. Bac Ninh which is considered \textbf{\textcolor{blue}{as a satellite city of Hanoi}}, there are many   industrial zones that bring high economic value to Vietnam, such as Samsung. }} &
  \textcolor{blue!80!black}{\textit{as a satellite city }} &
  \textcolor{blue!80!black}{\textit{a   satellite city}} \\ \cline{1-5}

  \vspace{1mm}
\multirow{2}{*}{\rotatebox[origin=c]{90}{\textbf{Why}}} &
  Vì sao khó xác định được lượng nước chính xác để trộn bột? &
  sử   dụng từ 200 cho tới 250 g nước trong các bạn thì à mà mình sử dụng nó sẽ \textbf{\textcolor{blue}{phụ   thuộc vào cái bột hút nước nhiều hay ít}} có nghĩa là nếu mà Bột mới thì nó sẽ   hút nước ít hơn là bột \underline{củ} và cái bột mới và \underline{một củ} thì các bạn sẽ tính vào   cái ngày sản xuất \vspace{1mm}&
  phụ   thuộc vào   bột hút nước nhiều hay ít &
  bột   hút nước nhiều hay ít \\
 &
  \textcolor{blue!80!black}{\textit{Why   is it hard to determine the exact amount of water to mix the dough?}} &
  \textcolor{blue!80!black}{\textit{The amount of   water you should use, between 200 and 250 grams, will \textbf{\textcolor{blue}{depend on the   absorbency of the flour}} which means The newer flour will be less absorbency than the \underline{old} one. The \underline{old} flour is determined by its production date.}} &
  \textcolor{blue!80!black}{\textit{depend   on the absorbency of the flour}} &
 \textcolor{blue!80!black}{\textit{ the   absorbency of the flour}} \\ \cline{1-5} 

\end{tabularx}

\label{fig:error}
\end{table}

\begin{table}[ht]
% \footnotesize
\begin{tabularx}{\textwidth}{|X|>{\centering\arraybackslash}p{0.18\textwidth}|p{0.45\textwidth}|p{0.1\textwidth}|p{0.1\textwidth}|}
    \hline
     & \textbf{Question} & \textbf{Transcript} & \textbf{Reference} & \textbf{Prediction} \\
    \hline
    \vspace{1mm}

  \vspace{1mm}
\multirow{2}{*}{\rotatebox[origin=c]{90}{\textbf{Where}}} &
  Món bánh hỏi nổi   tiếng nhất ở đâu? &
  cái   món bánh hỏi này nó có rất nhiều nơi nhưng để thành danh thì là \textbf{\textcolor{red}{mảnh đất võ}  \textcolor{blue}{ Bình Định nói chung và Quy Nhơn nói riêng}} nếu như chúng ta mà đi về đây mà   không thưởng thức món này thì có lẽ đó là một cái thiếu sót  \vspace{1mm} &
  Bình   Định nói chung và Quy Nhơn nói riêng &
  mảnh   đất võ Bình Định \\
 &
  \textcolor{blue!80!black}{\textit{Where is the   best place to try bánh hỏi? }} &
  \textcolor{blue!80!black}{\textit{This dish,   called 'bánh hỏi', is available in many places, but to taste the best, one   must visit \textbf{\textcolor{red}{the land of martial arts,} \textcolor{blue}{ Binh Dinh in general and Quy Nhon city   in particular}} If we come here and do not try this dish, it would be a   regrettable omission.}} &
  \textcolor{blue!80!black}{\textit{Binh   Dinh in general and Quy Nhon city in particular}} &
  \textcolor{blue!80!black}{\textit{the   land of martial arts, Binh Dinh}} \\ \cline{1-5} 
  \vspace{1mm}
\multirow{2}{*}{\rotatebox[origin=c]{90}{\textbf{When}}} &
  Lúc nào thì có   thể cho bánh vào dầu để chiên? &
  Bây   giờ thì mình sẽ đem đi \underline{chim} nha mình cho dầu ăn vào trong \underline{trở} và \underline{động cơ} nó   nóng lên nhé em \underline{rửa} và mình thấy \textbf{\textcolor{red}{cái} \textcolor{blue}{đầu nó sôi lăn tăn}} đây nè đó Mình sẽ cho   bánh vào nhá \vspace{1mm} &
  đầu   nó sôi lăn tăn &
  cái   đầu nó sôi lăn tăn \\
 & \textcolor{blue!80!black}{\textit{When cake should be put in the pan?}}
   &
  \textcolor{blue!80!black}{\textit{ I will \underline{fry} the cake in a moment, now  put the cooking oil to the \underline{pan} and \underline{wait for} it to be heated it up. Once it's hot, test it by dipping a \underline{chopstick} in; if   there \textbf{\textcolor{blue}{bubbles form around the tip}} it means the oil is ready. Then, I'll add   the cake to the pan.}} &
  \textcolor{blue!80!black}{\textit{bubbles   form around the tip}} &
  \textcolor{blue!80!black}{\textit{bubbles   form around the tip*}} \\ \cline{1-5} 
  \vspace{1mm}
\multirow{2}{*}{\rotatebox[origin=c]{90}{\textbf{Who}}} &
  Công thức được   chia sẻ này phù hợp với những ai? &
  nên   hôm nay \underline{là thay} chia sẻ cái công thức bột này \textbf{\textcolor{red}{tương đối dễ cho} \textcolor{blue}{các bạn mới   bắt đầu}} do đó là nếu mà các bạn cảm thấy là cái nguyên liệu này nó khó tìm   thì bạn có thể thay thế linh hoạt hơn thì vẫn cái bột vẫn chủ đạo nhất đó   chính là một mì  \vspace{1mm} &
  các   bạn mới bắt đầu &
  tương   đối dễ cho các bạn mới bắt đầu \\
 &
  \textcolor{blue!80!black}{\textit{Who does this recipe best suit?}} &
  \textcolor{blue!80!black}{\textit{So today,   \underline{Natha} share a flour recipe, \textbf{\textcolor{red}{relatively easy} \textcolor{blue}{for beginners}} If you find it   difficult to find the original ingredients, you can still be flexible and   replace them with other alternatives. The primary ingredient is still wheat flour.}} &
  \textcolor{blue!80!black}{\textit{beginners}} &
  \textcolor{blue!80!black}{\textit{relatively easy for beginners}} \\ \hline

\end{tabularx}
\caption{\centering Error examples for each question type of XLM-R model. The corresponding corrected English translations are also provided.}
\label{fig:error}
\end{table}

\end{document}